# Unmixing of Hyperspectral Data Using Robust Statistics-based NMF

Roozbeh Rajabi, Hassan Ghassemian
Electrical and Computer Engineering Department
Tarbiat Modares University
Tehran, Iran
r.rajabi@modares.ac.ir, ghassemi@modares.ac.ir

*Abstract*—Mixed pixels are presented in hyperspectral images due to low spatial resolution of hyperspectral sensors. Spectral unmixing decomposes mixed pixels spectra into endmembers spectra and abundance fractions. In this paper using of robust statistics-based nonnegative matrix factorization (RNMF) for spectral unmixing of hyperspectral data is investigated. RNMF uses a robust cost function and iterative updating procedure, so is not sensitive to outliers. This method has been applied to simulated data using USGS spectral library, AVIRIS and ROSIS datasets. Unmixing results are compared to traditional NMF method based on SAD and AAD measures. Results demonstrate that this method can be used efficiently for hyperspectral unmixing purposes.

## I. Introduction

Hyperspectral imaging has a key role in remote sensing applications. Hyperspectral sensors provide high spectral resolution that is useful for identification of materials presented in the scene. On the other hand the spatial resolution of hyperspectral images is low. This is due to technical restrictions of hyperspectral sensors.

As a result of low spatial resolution, hyperspectral images consist of mixed pixels. Mixed pixels are pixels containing more than one distinct material (see Fig. 1). Spectral unmixing algorithms try to decompose the observed spectra of these pixels into two set of information: Endmembers and abundance fractions (see Fig. 2) [1].

Spectral unmixing has been extensively studied during the last decade. Spectral unmixing algorithms can be categorized into geometrical and statistical categories [2]. Some of important and efficient methods for solving this problem are VCA [3], N-FINDR [4], NMF [5] and Sparse Methods [6].

Nonnegative matrix factorization (NMF) suits well to spectral unmixing problem, since endmembers and abundance fractions are nonnegative matrices. This method unlike other methods doesn't need the assumption of pure pixel presence.

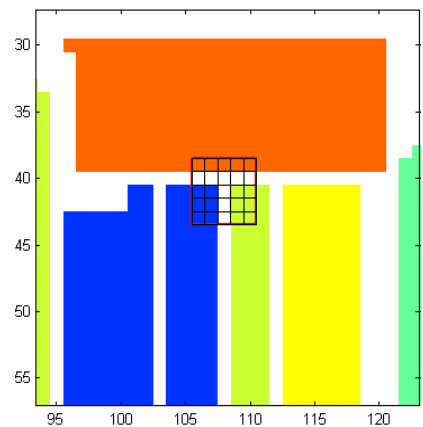

Figure 1. One mixed pixel selected in a scene and its subpixels.

There are extensions of NMF method proposed in the literature. For example CNMF [7] uses piecewise smoothness constraint. In this paper applicability of robust statistics-based nonnegative matrix factorization (RNMF) [8] for spectral unmixing has been investigated.

Section II introduces spectral unmixing problem and its mathematical formulation for linear model. Section III briefly reviews NMF and RNMF methods. In section IV experiments on various data has been done to evaluate the proposed method. Finally section V concludes the paper.

## II. Spectral Unmixing

### A. Problem definition

The problem of spectral unmixing refers to decomposition of observed spectra into endmembers spectra and abundance fractions. Endmembers are distinct materials that are presented in mixed pixels like asphalt, water, etc. Abundance fractions are the fractions in which endmembers appears in a mixed pixel [9]. Unmixing outputs are illustrated for a sample mixed pixel in Fig. 2.

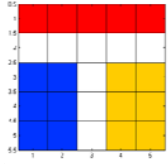

Extracted Endmembers:

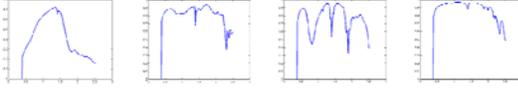

Abundance fractions:

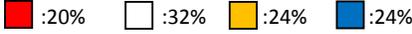

Figure 2. Spectral unmixing results for a sample mixed pixel

### B. Linear Mixture Model

In linear mixture model (LMM) it is assumed that the combination of spectral signatures is linear. Mathematical formulation of this model is expressed in (1).

$$X = AS + N \qquad (1)$$

Variables in (1) are summarized in Table I.

TABLE I.  VARIABLES IN LINEAR MIXTURE MODEL

| Variable | Description | Dimension |
|---|---|---|
| X | Observed data | M by L |
| A | Abundance fractions | M by P |
| S | Endmember signatures | P by L |
| N | Measurement noise | M by L |
| M | Total number of pixels | - |
| L | Number of spectral bands | - |
| P | Number of endmembers | - |

Two physical constraints exist on abundance fractions values. Firstly, abundance values are nonnegative, since they either exist in a mixed pixel (corresponding to positive abundance value) or does not exist (corresponding to zero abundance value). Secondly, sum of abundance values for a mixed pixel should be equal to one. These constraints are called ANC and ASC respectively [10].

## III. ROBUST STATISTICS-BASED NONNEGATIVE MATRIX FACTORIZATION

### A. Nonnegative Matrix factorization (NMF)

Nonnegative matrix factorization is a decomposition method that has been used in several applications. NMF is an unsupervised learning algorithm that decomposes a nonnegative matrix V into two nonnegative matrix factors W and H such that:

$$V \approx WH \qquad (2)$$

One of the cost functions to find factorization matrices can be defined based on Euclidean distance:

$$O(W, H) = \|V - WH\|^2 \qquad (3)$$

Multiplicative update rules can be used to minimize the cost function in (3) subject to the constraints $W, H \geq 0$ [11].

### B. Robust statistics-based Nonnegative Matrix Factorization (RNMF)

In this method another cost function different from regular NMF has been used. This cost function called hypersurface cost function is introduced by Samson et al. [12] and is defined as below:

$$\varphi(t) = \sqrt{1 + t^2} - 1 \qquad (4)$$

Hypersurface cost function is quadratic for small arguments and linear for large arguments as shown in Fig. 3. Another important characteristic of this function is differentiability of its influence function.

In RNMF method the cost function is as follows [8]:

$$\varphi(W, H) = \frac{1}{2}\left(\sqrt{1 + \|S - WH\|^2} - 1\right)$$
$$= \frac{1}{2}\left(\sqrt{1 + \sum_{ij}\left(S_{ij} - (WH)_{ij}\right)^2} - 1\right) \qquad (5)$$

### C. Updating Rules

Minimization on cost function in (5) yields the following iterative update rules [8]:

$$W_{ik}^{(t+1)} = W_{ik}^{(t)} - \alpha_{ik}^{(t)} \frac{(WHH^T)_{ik}^{(t)} - (SH^T)_{ik}^{(t)}}{\sqrt{1 + \|S - WH\|}}$$

$$H_{kj}^{(t+1)} = H_{kj}^{(t)} - \beta_{kj}^{(t)} \frac{(W^TWH)_{kj}^{(t)} - (W^TS)_{kj}^{(t)}}{\sqrt{1 + \|S - WH\|}} \qquad (6)$$

In these equations $\alpha_{ik}^{(t)}$ and $\beta_{kj}^{(t)}$ are the step sizes for each iteration t that are chosen via Armijo rule. Armijo rule is a method of line search that controls the step size for decreasing the cost function in a descent direction [13].

## IV. EXPERIMENTS AND RESULTS

Two experiments have been done for evaluation of the proposed method and results are compared with the results of original NMF method.

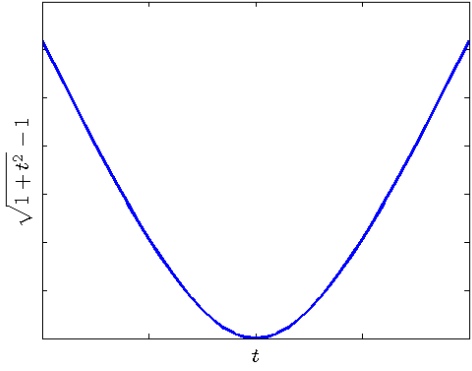

Figure 3. Hypersurface cost function

Simulated data has been used for these experiments. For generating simulated data, hyperspectral datasets and USGS spectral library [14] are used. First, reference spectral signatures from USGS are selected. The spectral signatures of each pixel in original dataset are replaced with these reference spectral signatures. Then resulted dataset has been filtered and down sampled to generate low spatial resolution data. Now the resulted data includes mixed pixels and can be used for evaluation of unmixing methods. Finally noise is added to data for simulating measurement noise. Algorithm for generating simulated data is illustrated in Fig. 4.

To evaluate unmixing methods different criteria are used in literature [15]. In this paper signature angle distance (SAD) and abundance angle distance (AAD) are used. These are defined in the following equations:

$$\text{SAD}_{m_i} = \cos^{-1}\left(\frac{m_i^T \hat{m}_i}{\|m_i\|\|\hat{m}_i\|}\right) \quad (7)$$

$$\text{AAD}_{a_i} = \cos^{-1}\left(\frac{a_i^T \hat{a}_i}{\|a_i\|\|\hat{a}_i\|}\right) \quad (8)$$

In (7) $m_i$ and $\hat{m}_i$ are i-th original signature and estimated one respectively. In (8) $a_i$ and $\hat{a}_i$ are i-th original abundance fractions and estimated one respectively. SAD and AAD measure the similarity between vectors. These measures do not depend on vector scales since they calculate angle between their input arguments.

The criteria in (7) and (8) defined for one endmember and one mixed pixel respectively. To obtain overall measure, RMS value of these criteria should be calculated.

$$\text{rms}_{\text{SAD}} = \sqrt{\frac{1}{p}\sum_{i=1}^{p}(\text{SAD}_{m_i})^2} \quad (9)$$

$$\text{rms}_{\text{AAD}} = \sqrt{\frac{1}{M}\sum_{i=1}^{M}(\text{AAD}_{a_i})^2} \quad (10)$$

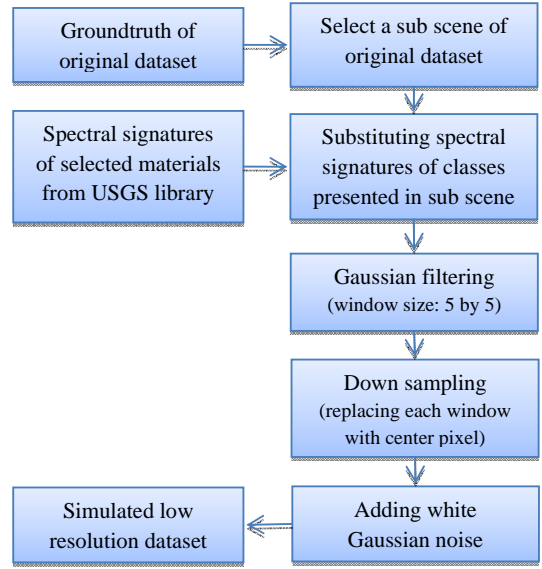

Figure 4. Algorithm for generating simulated data

### A. Experiment I

The first database that is used for evaluation is AVIRIS Indian Pines dataset [16]. The scene consists of 145 by 145 pixels and contains agriculture, forest and vegetation. Groundtruth map for this dataset is also available and illustrated in Fig. 5.

Table II summarizes implementation results of applying RNMF method on AVIRIS Indian Pines simulated dataset based on measures defined in (9) and (10).

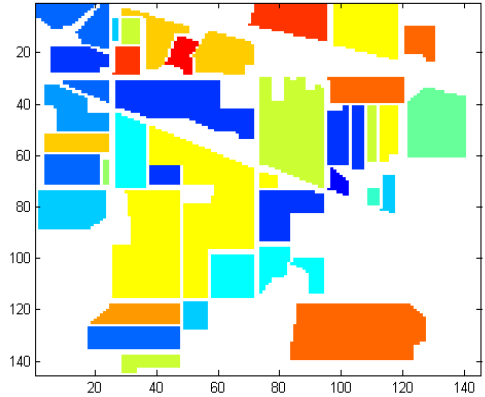

Figure 5. AVIRIS Indian Pines groundtruth map

TABLE II. RESULTS ON AVIRIS INDIANPINES SIMULATED DATA

|  | RMS value in degrees | |
| --- | --- | --- |
| Method | SAD | AAD |
| NMF | 17.26 | 25.06 |
| RNMF | 14.18 | 21.64 |

## B. Experiment II

Second experiment is done using ROSIS Pavia university dataset. The image size is 610 by 340 pixels. Groundtruth map of the dataset is shown in Fig. 6. This hyperspectral dataset is publicly available on the website of the Computational Intelligence Group from the Basque University (UPV/EHU) [17].

Table III summarizes implementation results of applying RNMF method on ROSIS Pavia University simulated dataset based on measures defined in (9) and (10).

TABLE III. RESULTS ON ROSIS PAVIA UNIVERSITY SIMULATED DATA

| Method | RMS value in degrees | |
|---|---|---|
|  | SAD | AAD |
| NMF | 15.78 | 23.12 |
| RNMF | 12.34 | 18.43 |

Results in Table II and III show that proposed method works better based on SAD and AAD measures.

## V. CONCLUSION AND REMARKS

Hyperspectral images contain mixed pixels, mainly because of low spatial resolution of the sensors. Spectral unmixing methods decompose a mixed pixel into endmembers and abundance fractions. In this paper robust statistics based nonnegative matrix factorization (RNMF) method has been used for spectral unmixing of hyperspectral images. This method due to using hypersurface cost function is robust against outliers. The proposed method is applied on simulated data and results are evaluated and compared against traditional NMF method. Simulated data are generated using AVIRIS Indian pines and Pavia University datasets. Spectral angle distance (SAD) and abundance angle distance (AAD) are used for comparing the methods. Results show that the proposed method based on RNMF can be used efficiently for hyperspectral unmixing purposes. Future works include adding more data specific criteria to the cost function and applying the method on real datasets rather than simulated ones.

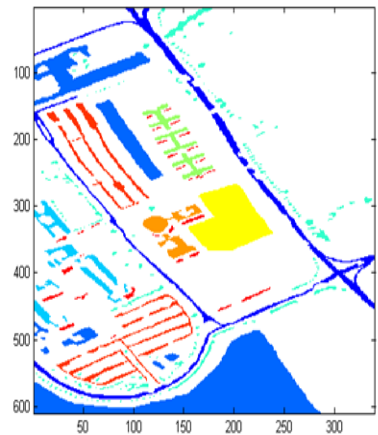

Figure 6. ROSIS Pavia University groundtruth map